\documentclass[table]{bmvc2k}
\usepackage{xcolor}
\usepackage{amsmath}
\usepackage{graphicx}
\usepackage{booktabs}
\usepackage{colortbl}
\usepackage{multirow}
\usepackage{multicol}
\usepackage{amssymb}
\usepackage{hhline}
\usepackage{booktabs}

\title{CPDR: Towards Highly-Efficient Salient Object Detection via Crossed Post-decoder Refinement}

\addauthor{Yijie Li}{yijieli2025@u.northwestern.edu}{1}
\addauthor{Hewei Wang}{heweiw@andrew.cmu.edu}{2}
\addauthor{Aggelos Katsaggelos}{a-katsaggelos@northwestern.edu}{1}

\addinstitution{
 Northwestern University\\
 633 Clark St \\
 Evanston, IL, USA
}
\addinstitution{
Robotics Institute\\
Carnegie Mellon University\\
Pittsburgh, PA, USA
}

\runninghead{Li, Wang, Katsaggelos}{CPDR}

\begin{document}

\maketitle

\begin{abstract}
Most of the current salient object detection approaches use deeper networks with large backbones to produce more accurate predictions, which results in a significant increase in computational complexity. A great number of network designs follow the pure UNet and Feature Pyramid Network (FPN) architecture which has limited feature extraction and aggregation ability which motivated us to design a lightweight post-decoder refinement module, the crossed post-decoder refinement (CPDR) to enhance the feature representation of a standard FPN or U-Net framework. Specifically, we introduce the Attention Down Sample Fusion (ADF), which employs channel attention mechanisms with attention maps generated by high-level representation to refine the low-level features, and Attention Up Sample Fusion (AUF), leveraging the low-level information to guide the high-level features through spatial attention. Additionally, we proposed the Dual Attention Cross Fusion (DACF) upon ADFs and AUFs, which reduces the number of parameters while maintaining the performance. Experiments on five benchmark datasets demonstrate that our method outperforms previous state-of-the-art approaches.
\end{abstract}

\section{Introduction}
Salient object detection (SOD) has rapidly evolved to become a cornerstone of modern computer vision, underpinning transformative advances across a diverse array of applications such as autonomous driving systems and robot exploration, where it enhances robotics real-time decision-making \cite{grigorescu2020survey, zhenqi23}, facilitates object manipulation and human-robot interaction \cite{garcia2015survey, wang2024airshot, zhu2023fanuc}. Additionally, SOD plays a crucial role in professional-grade image editing by simplifying complex object isolation tasks \cite{kumar2019survey, LIUCLOUDNET2024}. The significance of SOD lies in its ability to prioritize regions in visual scenes that most attract human attention, thus serving as a foundational technology for higher-level computer vision tasks including scene understanding and adaptive compression. Contemporary approaches in SOD predominantly harness deep convolutional neural networks, drawing on sophisticated architectural innovations that have shown remarkable success in feature representation and extraction. These methodologies often extend the foundational designs of U-Net \cite{ronneberger2015u} and Feature Pyramid Networks (FPN) \cite{zhao2019pyramid}, integrating multi-scale contextual information that significantly boosts the accuracy and robustness of saliency detection models.

However, despite these advances, there are several notable limitations in existing models:

\begin{itemize}
\item Deeper networks with large backbones and decoders significantly increase the computational burden, making them less feasible for deployment in resource-constrained environments.

\item The traditional FPN and U-Net architectures, though widely used, exhibit limited capability in optimal feature aggregation and representation, potentially compromising the detection performance in complex scenes.
\end{itemize}

To ameliorate the aforementioned issues, we are motivated to propose a novel lightweight architecture, the Crossed Post-decoder Refinement (CPDR), which introduces a highly-efficient post-decoder refinement module to enhance the feature representation in standard FPN or U-Net frameworks. Our approach strategically employs channel attention mechanisms to refine the low-level feature maps, which are then intricately crossed with spatial attention maps derived from high-level features. This cross-attention scheme ensures a more effective and efficient feature integration and representation.

The primary contributions of the CPDR architecture are threefold:

\begin{itemize}
\item We introduce a novel lightweight post-decoder refinement technique, Attention Down-Sample Fusion (ADF) and Attention Up-Sample Fusion (AUF) that integrates channel and spatial attention mechanisms to significantly enhance the saliency detection capabilities of traditional networks without special-designed decoders.

\item We propose a simplified CPDR module, Dual Attention Cross Fusion (DACF) with less computational complexity which is suitable for super-efficient salient object detection requirements, achieving 0.041 MAE with only 1.66M parameters.

\item Our proposed CPDR model demonstrates superior performance on five benchmark datasets, surpassing previous state-of-the-art methods while maintaining lower computational complexity.
\end{itemize}

Through these innovations, CPDR not only improves the effectiveness of salient object detection but also addresses the practical challenges associated with deploying deep learning models in environments with stringent resource constraints.

\section{Related Work}
\subsection*{Traditional Approaches in SOD}
Prior to the rise of deep learning, traditional SOD methods were predominantly reliant on heuristic features and hand-crafted models. Classic techniques such as contrast-based methods exploited the distinct visual features of objects to distinguish them from their surroundings \cite{itti1998model}. Region-based approaches took into account the homogeneity of regions to identify salient areas, often requiring significant feature engineering \cite{cheng2011global}. The most notable of these early attempts was the frequency-tuned method which utilized the color and luminance features to detect salient regions in an image, marking the beginning of more sophisticated SOD methodologies \cite{achanta2009frequency}. Methods like UFO, which stands for Uniqueness, Focusness, and Objectness, offered early insights into the core attributes that make objects stand out \cite{jiang2013salient}. A comprehensive benchmark by Borji et al. \cite{borji2015salient} provided a valuable evaluation of various traditional SOD algorithms, highlighting their strengths and limitations.

\begin{figure*}[htbp]
	\centering
	\includegraphics[width=5.0in]{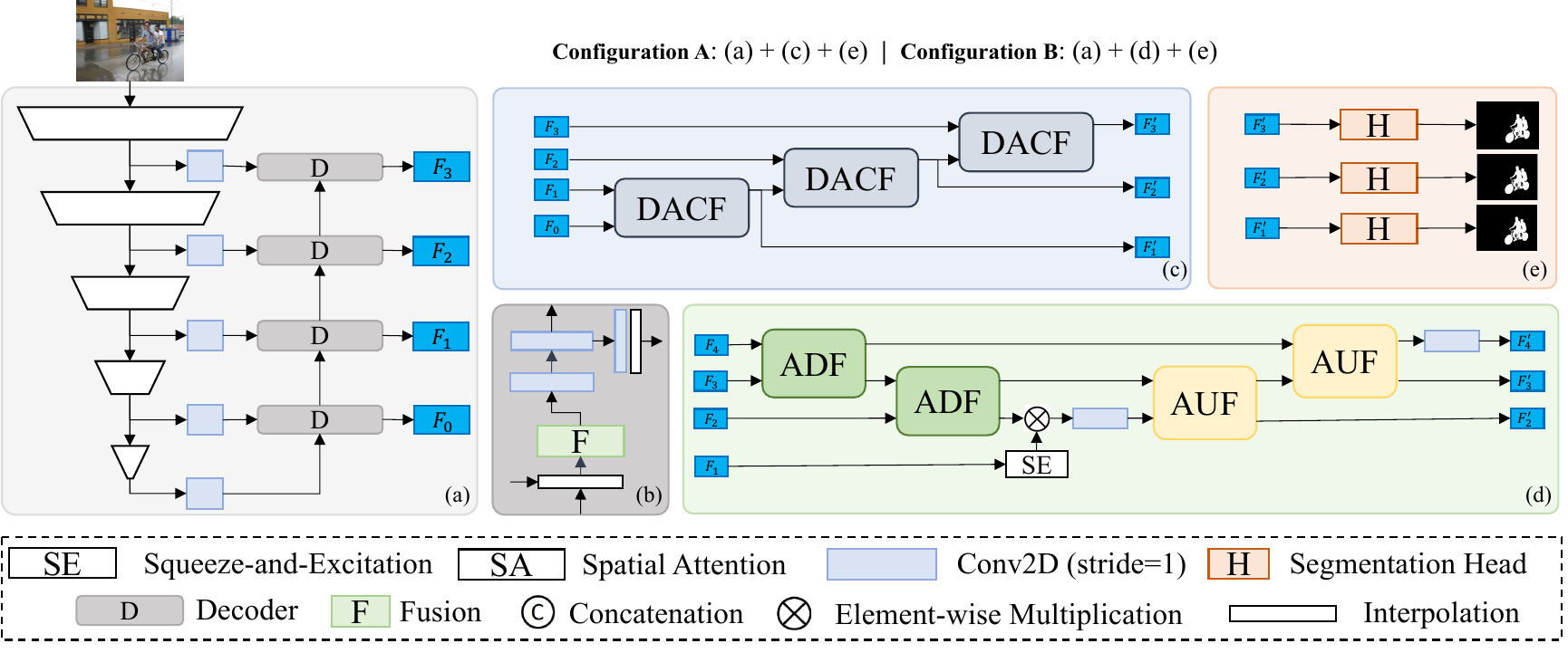}
	\caption{Overall Pipeline of CPDR}
	\label{fig:pipeline}
\end{figure*}

\subsection*{Large-Scale Learning-based Models in SOD}
As for the learning-based method, CNNs represent a prominent class of deep learning models that are extensively applied to the analysis and processing of 2D image data, demonstrating their effectiveness across a wide range of computer vision tasks~\cite{WANG2022102243, WANG2021, BATRA2022200039, WANGCAR2022}. The integration of attention mechanisms has further revolutionized the field by significantly enhancing the representation of visual features~\cite{HuoAtten, xu2024mentor, xu2024cikm}. In the realm of SOD, the emergence of deep learning has brought forth several large-scale models that stand out due to their intricate architectures and advanced feature extraction abilities. For instance, PoolNet leverages a multi-level feature aggregation strategy to refine saliency detection, offering impressive results across benchmark datasets \cite{liu2019poolnet}. Notably, recurrent strategies have been applied to SOD as well, as exemplified by Hu et al. \cite{hu2018recurrent}. Moreover, PiCANet proposed an innovative pixel-wise contextual attention mechanism \cite{liu2018picanet}, and Zhao et al. introduced the Pyramid Feature Attention Network (PFAN) \cite{zhao2019pyramid}, further enriching the landscape of SOD methodologies. BASNet introduces a boundary-aware inference mechanism that allows precise saliency mapping, often proving crucial for fine-grained object delineation \cite{qin2020basnet}. Similarly, U2Net emphasizes a dual attention mechanism that separates salient features from complex backgrounds, demonstrating remarkable adaptability to varied scenes \cite{zhang2021u2net}. C2SNet employs a contour-to-saliency transferring method that predicts salient objects and their contours simultaneously, enhancing detection accuracy \cite{wang2019c2snet}. Continuing to advance the field, PurNet, a novel approach, adopts a purity-oriented structure to isolate salient objects against noisy backgrounds \cite{zhou2020purnet}. Lastly, Dual Attention Aggregating Network (DAANet) introduces a dual attention aggregating network, adding a new layer of feature refinement and aggregation for enhanced saliency detection \cite{li2023daanet}.

\subsection*{Lightweight Learning-based Models in SOD}
With the increasing demand for real-time applications, there has been a focus on developing lightweight models that maintain a balance between efficiency and accuracy. The EDN model developed by Wu et al. in 2022, which stands for Extremely-Downsampled Network (EDN), showcases a high level of precision through novel feature compression techniques that do not compromise detection quality \cite{wu2022edn}. This model is particularly suitable for deployment in resource-constrained environments where computational efficiency is as vital as the accuracy of the saliency detection. In contrast, the Efficient Detection Network (EDN) model from earlier works like Mei et al. in 2017 refers to a different approach and should be clearly differentiated to avoid confusion \cite{mei2017edn}. The Cross-Attention Siamese Network (CASNet) focuses on video SOD by efficiently processing temporal information across frames \cite{luo2020casnet}. Additionally, the Hierarchical Visual Perception Network (HVPNet) \cite{liu2020lightweight} and Stereoscopically Attentive Multi-scale Network (SAMNet) \cite{liu2021samnet} enhance lightweight saliency detection by utilizing hierarchical and attentive multi-scale processing. The CII18 model, which rethinks traditional U-shape structures, further improves performance and efficiency in saliency detection \cite{liu2021rethinking}. These models, including HVPNet, SAMNet, CII18, and the two distinct EDN models, exemplify the trend toward solutions that are both computationally efficient and highly accurate in detecting salient objects.

\section{Crossed Post-decoder Refinement (CPDR)}
\subsection{Overview of CPDR}
As shown in Figure \ref{fig:pipeline}, our proposed Crossed Post-decoder Refinement (CPDR) serves as a refinement module after the decoders. In Figure \ref{fig:pipeline} (a), we use standard UNet \cite{ronneberger2015u} and Feature Pyramid Network (FPN) \cite{lin2017feature} with modified MobileNetV2 \cite{sandler2018mobilenetv2} and EfficientNet \cite{tan2019efficientnet} B0/B3 backbones for experiments. The structure of the decoders is shown in Figure \ref{fig:pipeline} (b) which only contains three convolution layers without special design. In our smallest version (Ours-S), we modified the original MobileNetV2 \cite{sandler2018mobilenetv2} by removing the last two blocks which gives us a simplified backbone with only 1.4 million parameters. In this configuration, we use FPN as the encoder-decoder network, with a Dual Attention Cross Fusion (DACF) as the refinement module, shown in Figure \ref{fig:pipeline} (c). As for our medium (Ours-M) and large version (Ours-L), we use EfficientNet-B0 and EfficientNet-B3 as backbone, UNet as encoder-decoder network, and their refinement module is a U-shaped module, illustrated in Figure \ref{fig:pipeline} (d).

\begin{figure*}[t]
	\centering
	\includegraphics[width=5.0in]{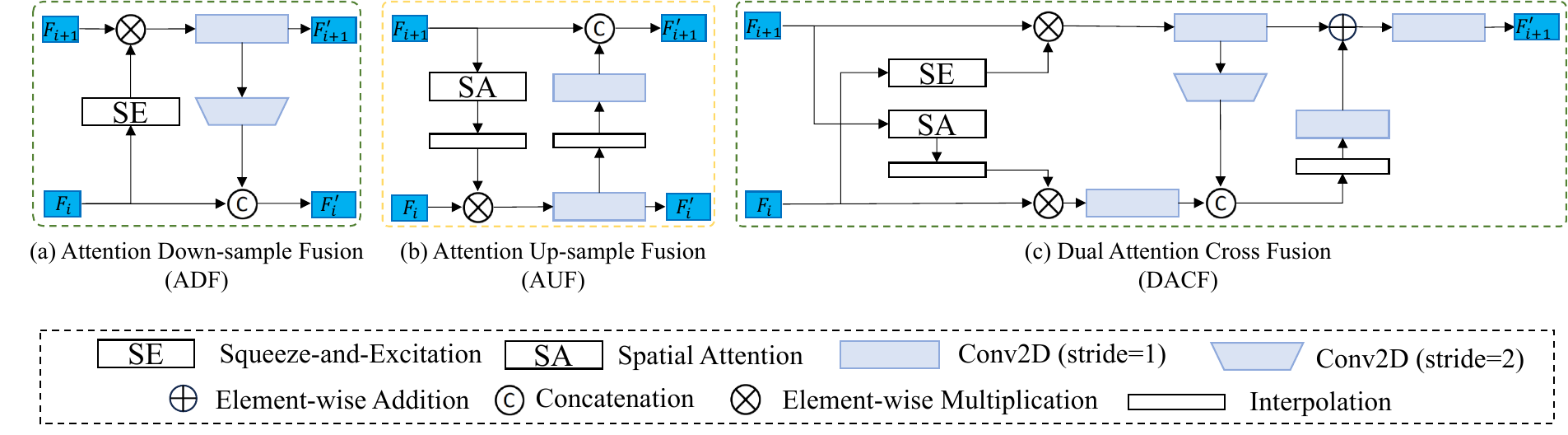}
	\caption{Demonstration of CPDR Modules}
	\label{fig:modules}
\end{figure*}

\subsection{Attention Down-Sample Fusion (ADF)}
The channels attention mechanism is widely used in computer vision tasks, which emphasize certain channels over others. However, most of the current channels attention is applied to the same feature maps, focusing on single-level feature re-weight. Motivated by Squeeze-and-Excitation (SE) \cite{hu2018squeeze}, we proposed the Attention Down-Sample Fusion (ADF) with cross-level channels attention for more effective feature fusion, shown in Figure \ref{fig:modules} (a). The basic idea of ADF is the channel dimension of a high-level feature consists of richer information, which can be utilized to create an attention map for lower-level features. Considering a high-level feature $F_{i}$ and a low-level feature $F_{i+1}$, we compute the $F^{'}_{i+1}$ by applying the channels attention followed by a convolution, while $F^{'}_{i}$ can be calculated based on $F^{'}_{i+1}$ output:

\begin{equation}
    F^{'}_{i+1} = \mathrm{Conv2D}(\mathrm{SE}(F_{i}) \odot F_{i+1})
\end{equation}

\begin{equation}
    F^{'}_{i} = \mathrm{Concat(\{F_{i}, \mathrm{Down(F^{'}_{i+1})}\})}
\end{equation}
    
\noindent where $\mathrm{Down}$ is a convolution layer with a stride equal to 2. The ADF provided a simple way to effectively fuse the low-level feature with high-level features under a `cross-attention' manner.

\subsection{Attention Up-Sample Fusion (AUF)}
However, as shown in Figure \ref{fig:pipeline} (d), we only consider $F^{'}_{3}$ as the final output, which means features fused by ADFs don't have any contribution to the final output, so we proposed the Attention Up-Sample Fusion (AUF) to match these ADFs, shown in Figure \ref{fig:modules} (b). Sharing a similar idea of ADFs, AUF considers that lower-level feature with higher resolution contains more accurate spatial information which can be used to create attention for high-level features. Then, the output $F^{'}_{i}$ can be calculated by:

\begin{equation}
    \mathrm{Conv2D}(\mathrm{Interpolate}(\mathrm{SA}(F_{i+1})) \odot F_{i})
\end{equation}

\noindent where $\mathrm{SA}$ is the CBAM \cite{woo2018cbam} spatial attention and $\mathrm{Interpolate}$ represent the bilinear interpolation. The output $F^{'}_{i+1}$ can be computed through:

\begin{equation}
    \mathrm{Concat}(\{F_{i+1}, \mathrm{Conv2D}(\mathrm{Interpolate}(F^{'}_{i})\}))
\end{equation}

\noindent The AUFs successfully fused all feature maps generated by ADFs to the final output by utilizing the guidance of spatial attention.

\subsection{Dual Attention Cross Fusion (DACF)}
In order to further reduce the computational complexity, we introduce Dual Attention Cross Fusion (DACF), which combines an ADF and an AUF with some of the $3 \times 3$ convolution layers replaced by $1 \times 1$ convolution layers, as shown in Figure \ref{fig:modules} (c). The DACF separately applies channel attention and spatial attention to low-level feature maps and high-level feature maps, which is different from the ADF and AUF pipeline. The weighted $F_{i+1}$ ($wF_{i+1}$) and $F_{i}$ ($wF_{i}$) can be directly calculated by the same method used in ADF and AUF. As for the fusion, we first concatenate the two weighted feature maps though $\mathrm{Concat}(\{\mathrm{Down}(wF_{i+1}), wF_{i}\})$ to produce $C_{i}$. Finally, we aggregate $C_{i}$ to generate output $F^{'}_{i+1}$ by computing:

\begin{equation}
    \mathrm{Conv2D}(\mathrm{Conv2D}(\mathrm{interpolate}(C_{i}))+wF_{i+1})
\end{equation}

\noindent The DACF simplified the ADF+AUF pipeline with less computation complexity which makes DACF suitable for the small configuration, Ours-S.

\subsection{Objective Functions}
We use DICE loss and Intersection over Union (IoU) loss for training. The DICE loss can be formulated as:

\begin{equation}
    \mathcal{L}_{\mathrm{DICE}} = 1 - \frac{\sum_{m, n}p \odot y + \epsilon}{\sum_{m, n}p + \sum_{m, n}y + \epsilon}
\end{equation}

\noindent and IOU loss is:

\begin{equation}
    \mathcal{L}_{\mathrm{IOU}} = 1 - \frac{\sum_{m, n}p \odot y + \epsilon}{\sum_{m, n}p + \sum_{m, n}y - \sum_{m, n}p \odot y + \epsilon}
\end{equation}

\noindent where $p$ represent the prediction and $y$ represent the ground truth. $\epsilon$ indicate the smoothing coefficient and $\odot$ is the element-wise multiplication. Then, the total loss functions can be formulated by:

\begin{equation}
    \mathcal{L}_{\mathrm{total}} = \mathcal{L}_{\mathrm{DICE}} + \mathcal{L}_{\mathrm{IOU}}
\end{equation}

\noindent In order to speed up the convergence speed, we use the deep-supervision strategy which means the $\mathcal{L}_{\mathrm{total}}$ will be applied to the three predictions from three stages and ground truth will be interpolated to match the size of each prediction.

\section{Experiments}
\subsection{Datasets} \label{dataset_section}
We utilize the DUTS-TR dataset \cite{wang2017learning} to train our DAANet, assessing its performance on multiple benchmarks including DUTS-TE \cite{wang2017learning}, HKU-IS \cite{li2015visual}, ECSSD \cite{yan2013hierarchical}, PASCAL-S \cite{li2014secrets}, and DUT-OMRON \cite{yang2013saliency}. The training dataset, DUTS-TR, comprises 10,553 samples. For evaluation, DUTS-TE provides 5,019 images, which is the most widely-used test set for SOD task. HKU-IS features 4,447 image pairs, and ECSSD offers 1,000 test images. Additionally, PASCAL-S contains 850 images and DUT-OMRON includes 5,168 images for testing.

\subsection{Evaluation Metrics} \label{metrics_section}
We employ five metrics to assess the performance of CPDR: 

\begin{itemize}
    \item Mean Absolute Error (MAE) \cite{perazzi2012saliency}, defined as $\frac{1}{N \times M} \sum_{n=1}^{N} \sum_{m=1}^{M} |t_m - p_m|$, which quantifies the average per-pixel discrepancy between the prediction and ground truth.
    \item Mean F-measure ($F^{m}_{\beta}$), serves as the weighted harmonic mean between precision and recall, which is formulated by $\frac{(1 + \beta^2) \cdot \text{Precision} \cdot \text{Recall}}{\beta^2 \cdot \text{Precision} + \text{Recall}}$, where $\beta^2 = 0.3$.
    \item Mean E-measure ($E^{m}_{\phi}$) \cite{fan2021cognitive}, combines the single pixel values with the global-level mean value, which can be calculated as $\frac{1}{N \times M} \sum_{n=1}^{N} \sum_{m=1}^{M} \theta(\phi)$, where $\phi$ is the alignment matrix and $\theta(\phi)$ is the enhanced alignment matrix.
    \item $S$-measure $(S_m)$ \cite{Fan_2017_ICCV}, designed to quantitatively evaluate structural similarity in a manner that closely aligns with human visual perception. This metric is meticulously computed using the formula $S_m = m \cdot s_o + (1 - m) \cdot s_r $, where $s_o$ and $s_r$ denote the object-aware and region-aware structural similarities, respectively. The parameter $m$ is strategically set to 0.5 to balance the influence of both components.
    \item Weighted F-measure $(F_\beta^\omega)$ \cite{fan2021cognitive} offers a nuanced adaptation of the traditional $F_\beta$ measure by emphasizing precision and recall differently based on the spatial characteristics of the errors. It is calculated as:
\[
F_\beta^\omega = \frac{(1+\beta^2) \cdot (\omega_p \cdot \text{Precision}) \cdot (\omega_r \cdot \text{Recall})}
{\beta^2 \cdot (\omega_p \cdot \text{Precision}) + (\omega_r \cdot \text{Recall})},
\]

where $\omega_p$ and $\omega_r$ are weighting functions that adaptively modify the influence of precision and recall, respectively, based on localized error characteristics and $\beta^2 = 0.3$. This metric considers the specific locations and neighborhood context, making it especially suitable for non-binary evaluation scenarios.

\end{itemize}

\subsection{Implementation Detail}
We perform the training on a single NVIDIA Tesla V100-SXM2 16GB GPU. We train our models on DUTS-TR \cite{wang2017learning} dataset. The training batch size for all experiments is set to 16 and 40 epochs for the model with MobileNetV2 backbone and 20 epochs with EfficientNet backbones. We use Adam as the training optimizer. A poly learning rate scheduler with a linear warm-up is adopted, we warm up the training with 5 epochs, and $\gamma$ for poly learning rate decay is set to 3 for MobileNetV2 backbone and 5 for EfficientNet backbones. Images are resized to $(256, 256)$ and only a random horizontal flip is adopted for data augmentation.

\subsection{Ablation Study}

\begin{table*}[htbp]
	\centering
        \caption{Ablation study on module compositions, loss functions, and backbone networks}
	\scalebox{0.65}{
    	\begin{tabular}{ccccccccccccccc}
    		\cline{1-15}
    		\multicolumn{1}{c|}{\multirow{2}{*}{\textbf{No.}}} & \multicolumn{1}{c|}{\multirow{2}{*}{\textbf{Backbone}}} & \multicolumn{1}{c|}{\multirow{2}{*}{\textbf{Arch}}} & \multicolumn{3}{c|}{\textbf{CPDR Config}} & \multicolumn{3}{c|}{\textbf{Loss Function}} & \multicolumn{1}{c|}{\textbf{\#Params}} & \multicolumn{5}{c}{\textbf{DUTS-TE}} \\
    		\multicolumn{1}{c|}{} & \multicolumn{1}{c|}{} & \multicolumn{1}{c|}{} & ADF & AUF & \multicolumn{1}{c|}{DACF} & DICE & BCE & \multicolumn{1}{c|}{IOU} & \multicolumn{1}{c|}{(M)} & $\mathcal{M\downarrow}$ & $F_{\beta}^{m}\uparrow$ & $F_{\beta}^{\omega}\uparrow$ & $S_{m}\uparrow$ & $E_{\phi}^{m}\uparrow$ \\ 
    		\cline{1-15}
                \multicolumn{1}{c|}{1} & \multicolumn{1}{c|}{\multirow{6}{*}{MobileNetV2*}} & \multicolumn{1}{c|}{FPN} &  &  &  &  & $\checkmark$ & \multicolumn{1}{c|}{$\checkmark$} & \multicolumn{1}{c|}{1.58} & .048 & .800 & .776 & .850 & .899\\
                \multicolumn{1}{c|}{2} & \multicolumn{1}{c|}{} & \multicolumn{1}{c|}{FPN} & $\checkmark$ & $\checkmark$ &  &  & $\checkmark$ & \multicolumn{1}{c|}{$\checkmark$} & \multicolumn{1}{c|}{1.72} & .044 & .812 & .790 & .856 & .903\\
                \multicolumn{1}{c|}{3} & \multicolumn{1}{c|}{} & \multicolumn{1}{c|}{UNet} & $\checkmark$ & $\checkmark$ &  &  & $\checkmark$ & \multicolumn{1}{c|}{$\checkmark$} & \multicolumn{1}{c|}{1.82} & .044 & .814 & .793 & .857 & .905\\
                \multicolumn{1}{c|}{4} & \multicolumn{1}{c|}{} & \multicolumn{1}{c|}{UNet} &  &  & $\checkmark$ &  & $\checkmark$ & \multicolumn{1}{c|}{$\checkmark$} & \multicolumn{1}{c|}{1.75} & .044 & .813 & .791 & .855 & .904\\
                \multicolumn{1}{c|}{5} & \multicolumn{1}{c|}{} & \multicolumn{1}{c|}{FPN} &  &  & $\checkmark$ &  & $\checkmark$ & \multicolumn{1}{c|}{$\checkmark$} & \multicolumn{1}{c|}{1.66} & .044 & .810 & .787 & .854 & .902\\
                \rowcolor{gray!25}
                \multicolumn{1}{c|}{6} & \multicolumn{1}{c|}{} & \multicolumn{1}{c|}{FPN} &  &  & $\checkmark$ & $\checkmark$ &  & \multicolumn{1}{c|}{$\checkmark$} & \multicolumn{1}{c|}{1.66} & .041 & .822 & .803 & .851 & .910\\
    		\cline{1-15}
    		\multicolumn{1}{c|}{7} & \multicolumn{1}{c|}{\multirow{3}{*}{EfficientNet-B0}} & \multicolumn{1}{c|}{FPN} & $\checkmark$ & $\checkmark$ &  & & $\checkmark$ & \multicolumn{1}{c|}{$\checkmark$} & \multicolumn{1}{c|}{4.54} & .041 & .826 & .808 & .867 & .912\\
                \multicolumn{1}{c|}{8} & \multicolumn{1}{c|}{} & \multicolumn{1}{c|}{UNet} & $\checkmark$ & $\checkmark$ &  &  & $\checkmark$ & \multicolumn{1}{c|}{$\checkmark$} & \multicolumn{1}{c|}{4.64} & .040 & .828 & .811 & .869 & .912\\
                \rowcolor{gray!25}
                \multicolumn{1}{c|}{9} & \multicolumn{1}{c|}{} & \multicolumn{1}{c|}{UNet} & $\checkmark$ & $\checkmark$ &  & $\checkmark$ &  & \multicolumn{1}{c|}{$\checkmark$} & \multicolumn{1}{c|}{4.64} & .038 & .839 & .825 & .864 & .922\\
                \arrayrulecolor{black}\hhline{---------------}
                \rowcolor{gray!25}
                \multicolumn{1}{c|}{10} & \multicolumn{1}{c|}{EffcientNet-B3} & \multicolumn{1}{c|}{UNet} & $\checkmark$ & $\checkmark$ &  & $\checkmark$ &  & \multicolumn{1}{c|}{$\checkmark$} & \multicolumn{1}{c|}{11.36} & \textbf{.034} & \textbf{.853} & \textbf{.842} & \textbf{.875} & \textbf{.931}\\
                \cline{1-15}
    	\end{tabular}
	}
	\label{tab:ablation_result}
\end{table*}

\noindent To demonstrate the effectiveness of our Crossed Post-decoder Refinement (CPDR) module, we conduct a comprehensive ablation study on the backbone networks, encoder-decoder architectures, module compositions, and loss functions, as shown in Table \ref{tab:ablation_result} on DUTS-TE \cite{wang2017learning} dataset with five metrics: MAE ($\mathcal{M}$), mean F-measure ($F^{m}_{\beta}$), weighted F-measure ($F^{\omega}_{\beta}$), S-measure ($S_{m}$), and mean E-measure ($E^{m}_{\phi}$). We consider an FPN architecture with our modified MobileNetV2 \cite{sandler2018mobilenetv2} backbone trained with BCE and IOU loss as the baseline (No. 1), which has 1.58M parameters. It can achieve an MAE of 0.048 and a weighted F-measure of 0.8. In experiment No. 2, we maintain the FPN structure and equip the pipeline with ADFs and AUFs, with only a 0.14M increase in parameters amount, but the performance on all five metrics is significantly improved with a new MAE of 0.044. By comparing No. 2 and No. 3, we find that adopting UNet \cite{ronneberger2015u} instead of FPN \cite{zhao2019pyramid} will increase the number of parameters but the performance improvement is not significant. In experiments No. 4 and No. 5, we compare the performance of UNet and FPN structure with DACF, and we observe that the parameters amount is reduced without significant performance drop. In No. 6, we adopt DICE+IOU as loss functions. Considering many previous approaches have demonstrated the effectiveness of IOU loss, we only compare the results of No. 6 with No. 5 (BCE+IOU) which shows that DICE+IOU is better than BCE+IOU which gets an MAE of 0.041 with only 1.66M parameters using FPN and DACF. From No. 7 to No. 9, we conduct experiments using EfficientNet-B0 backbone \cite{tan2019efficientnet}, the results illustrate that UNet structure with ADF and AUF trained with DICE and IOU loss can achieve the best performance. Finally, in No. 10, utilizing the EfficientNet-B3 backbone, our approach gets an MAE of 0.034 with only 11M parameters.

\subsection{Qualitative Evaluation}
\begin{figure*}[t]
	\centering
	\includegraphics[width=5.0in]{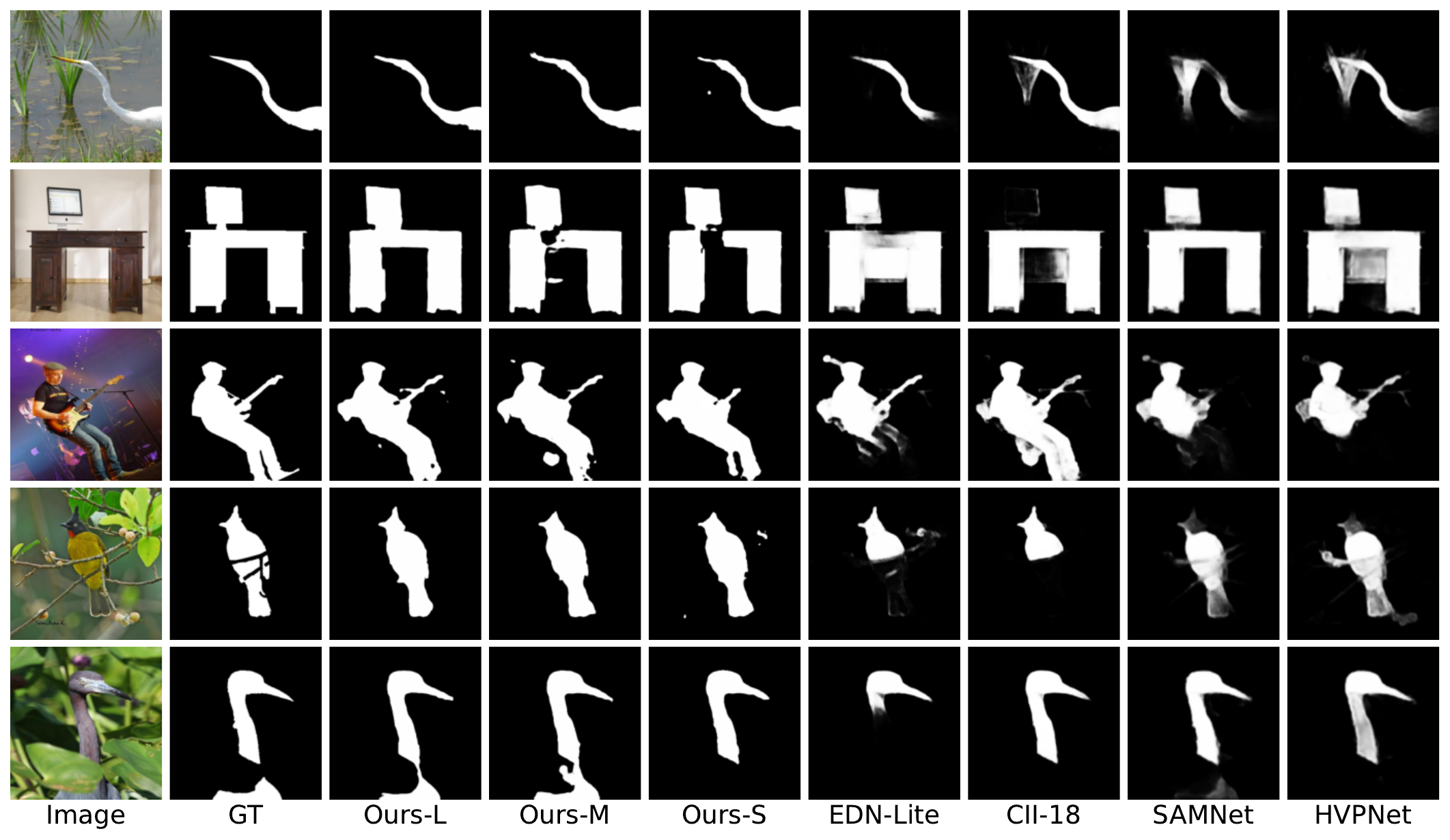}
	\caption{Qualitative comparison on DUTS-TE dataset between our methods and previous lightweight state-of-the-art approaches}
	\label{res_compare_qua}
\end{figure*}

Figure \ref{res_compare_qua} illustrates the qualitative results on DUTS-TE dataset. We can observe that our model (Ours-L, Ours-M, and Ours-S), achieves superior segmentation performance and completeness. Our models adeptly handle diverse and challenging imaging conditions. They excel in scenarios involving complex structures in row 2 and 4, multiple objects such as row 3, where other models often lose detail or misinterpret the boundaries, and cluttered backgrounds like row 5, as well as in distinguishing small and intricate details such as those seen in rows 4 and 5. These results demonstrate the capability of our CPDR models to deliver high-quality segmentation outputs consistently across various challenging scenarios. 

\subsection{Quantitative Evaluation}

\begin{figure*}[htbp]
    \centering

    \subfigure{
        \includegraphics[width=0.2\textwidth]{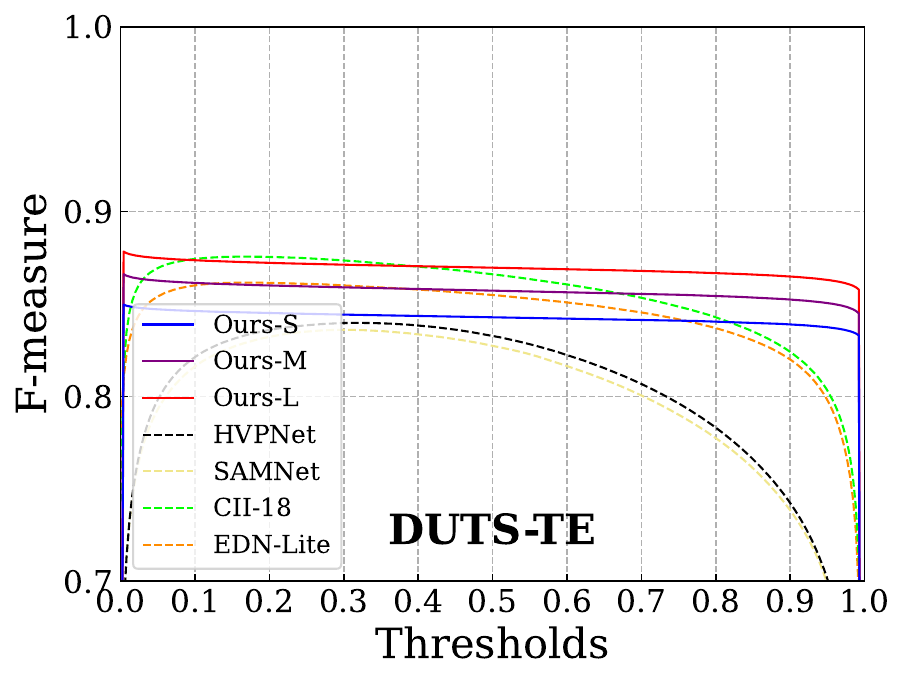}}
    \hspace{-3mm}
    \subfigure{
        \includegraphics[width=0.2\textwidth]{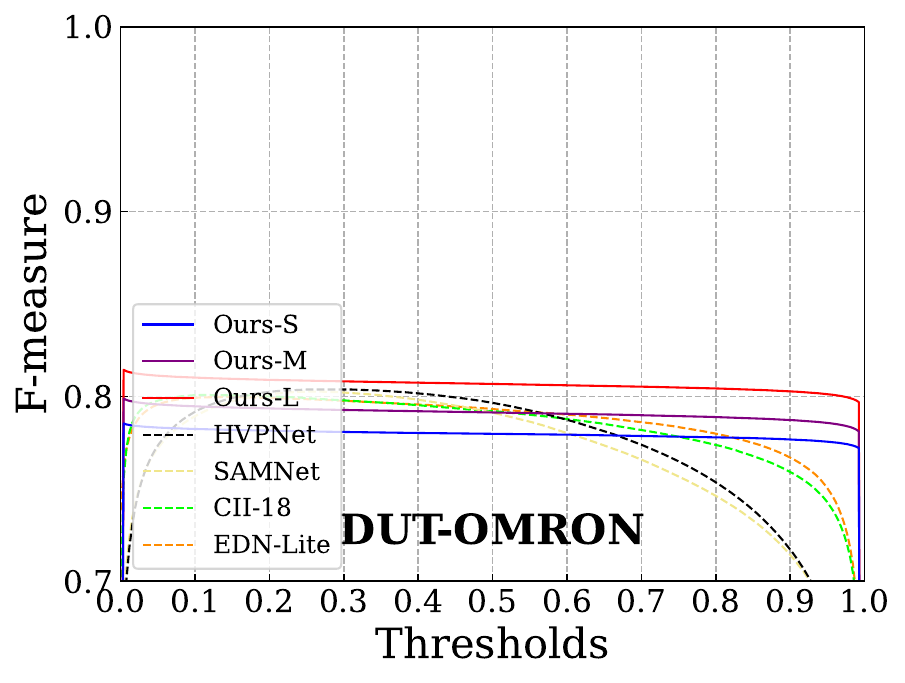}}
    \hspace{-3mm}
    \subfigure{
        \includegraphics[width=0.2\textwidth]{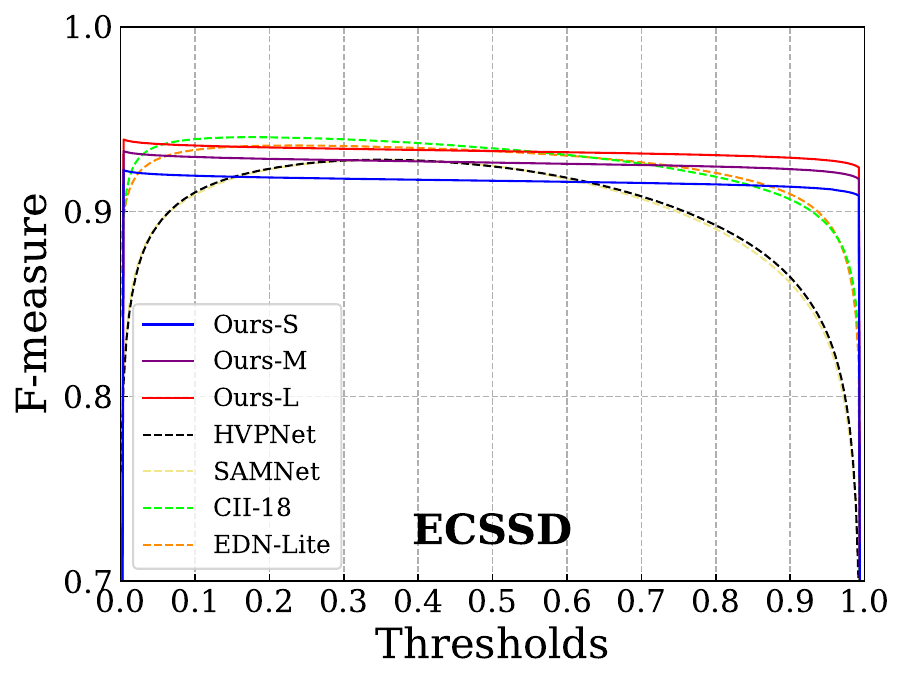}}
    \hspace{-3mm}
    \subfigure{
        \includegraphics[width=0.2\textwidth]{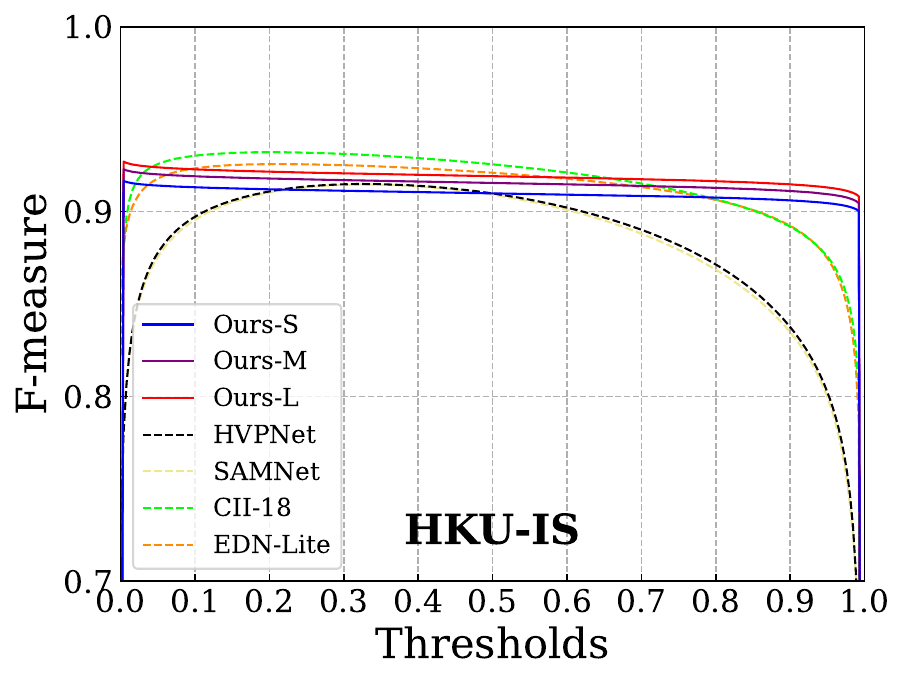}}
    \hspace{-3mm}
    \subfigure{
        \includegraphics[width=0.2\textwidth]{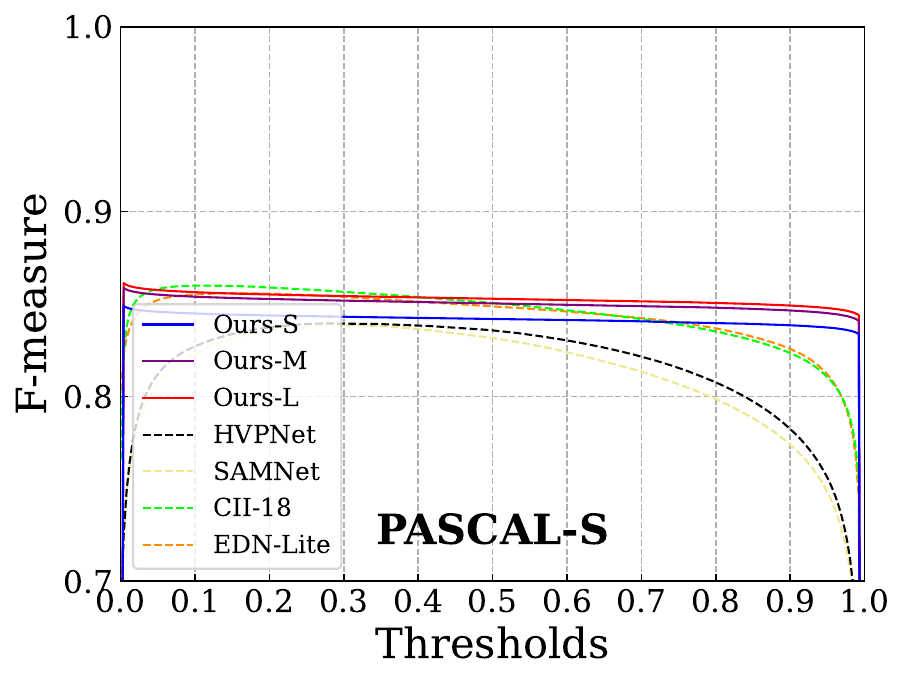}}
    \caption{Illustration of F-measure curves on five benchmark datasets}
    \label{pr_fm_curves}
\end{figure*}

Table \ref{quan_result} and F-measure curves shown in Figure \ref{pr_fm_curves} provide a comprehensive quantitative comparison and performance visualization of various SOD models across multiple datasets. Notably, three versions of a new model, designated as Ours (S), Ours (M), and Ours (L), demonstrate varying levels of efficiency and effectiveness. Ours (S) achieves significant compactness with only 1.66M parameters and 1.02 MACs, yet maintains competitive performance metrics. Ours (M) strikes an optimal balance between speed and accuracy with its 4.64M parameters, outperforming many existing models. Ours (L) excels in performance, utilizing 11.36M parameters to surpass nearly all advanced SOD methods, thus establishing new benchmarks in both efficiency and effectiveness. 

\begin{table*}[htbp]
	\centering
        \caption{Quantitative comparison with state-of-the-art approaches}
	\scalebox{0.54}{
    	\begin{tabular}{cccccccccccccccccc}
    		\cline{1-18}
    		\multicolumn{1}{c|}{\multirow{2}{*}{\textbf{Methods}}} & \multicolumn{1}{c|}{\textbf{\#Params}} & \multicolumn{1}{c|}{\textbf{MACs}} & \multicolumn{3}{c|}{\textbf{ECSSD}}  & \multicolumn{3}{c|}{\textbf{PASCAL-S}} & \multicolumn{3}{c|}{\textbf{DUTS-TE}} & \multicolumn{3}{c|}{\textbf{HKU-IS}} & \multicolumn{3}{c}{\textbf{DUT-OMRON}}\\
    		\multicolumn{1}{c|}{} & \multicolumn{1}{c|}{(M)} & \multicolumn{1}{c|}{(G)} & $\mathcal{M\downarrow}$ & $F_{\beta}^{m}\uparrow$ & \multicolumn{1}{c|}{$E_{\phi}^{m}\uparrow$} & $\mathcal{M\downarrow}$ & $F_{\beta}^{m}\uparrow$ & \multicolumn{1}{c|}{$E_{\phi}^{m}\uparrow$} & $\mathcal{M\downarrow}$ & $F_{\beta}^{m}\uparrow$ & \multicolumn{1}{c|}{$E_{\phi}^{m}\uparrow$} & $\mathcal{M\downarrow}$ & $F_{\beta}^{m}\uparrow$ & \multicolumn{1}{c|}{$E_{\phi}^{m}\uparrow$} & $\mathcal{M\downarrow}$ & $F_{\beta}^{m}\uparrow$ & $E_{\phi}^{m}\uparrow$ \\ 
    		\cline{1-18}
                \multicolumn{18}{c}{Conventional CNN Models} \\
                \cline{1-18}
                \multicolumn{1}{c|}{$\mathrm{BASNet_{19}}$ \cite{qin2019basnet}} & \multicolumn{1}{c|}{87.06} & \multicolumn{1}{c|}{127.36} & .037 & .917 & \multicolumn{1}{c|}{.943} & .076 & .818 & \multicolumn{1}{c|}{.879} & .048 & .823 & \multicolumn{1}{c|}{.896} & .032 & .902 & \multicolumn{1}{c|}{.943} & .056 & .767 & .865\\
                \multicolumn{1}{c|}{$\mathrm{MINet_{20}}$ \cite{pang2020multi}} & \multicolumn{1}{c|}{126.38} & \multicolumn{1}{c|}{87.11} & .033 & .923 & \multicolumn{1}{c|}{.950} & .064 & .830 & \multicolumn{1}{c|}{.896} & .037 & .844 & \multicolumn{1}{c|}{.917} & .029 & .909 & \multicolumn{1}{c|}{.952} & .056 & .757 & .860\\
                \multicolumn{1}{c|}{$\mathrm{GateNet_{20}}$ \cite{zhao2020suppress}} & \multicolumn{1}{c|}{128.63} & \multicolumn{1}{c|}{162.13} & .033 & .927 & \multicolumn{1}{c|}{.948} & .062 & .844 & \multicolumn{1}{c|}{.901} & .037 & .851 & \multicolumn{1}{c|}{.917} & .029 & .916 & \multicolumn{1}{c|}{.952} & .054 & .770 & .865\\
                \multicolumn{1}{c|}{$\mathrm{CII50_{21}}$ \cite{liu2021rethinking}} & \multicolumn{1}{c|}{24.48} & \multicolumn{1}{c|}{11.60} & .033 & .927 & \multicolumn{1}{c|}{.948} & .062 & .844 & \multicolumn{1}{c|}{.901} & .037 & .851 & \multicolumn{1}{c|}{.917} & .029 & .916 & \multicolumn{1}{c|}{.952} & \textbf{.054} & .770 & .865\\
                \multicolumn{1}{c|}{$\mathrm{EDN_{22}}$ \cite{wu2022edn}} & \multicolumn{1}{c|}{42.85} & \multicolumn{1}{c|}{20.45} & .032 & \textbf{.930} & \multicolumn{1}{c|}{.951} & .062 & \textbf{.849} & \multicolumn{1}{c|}{.902} & .035 & .863 & \multicolumn{1}{c|}{.925} & .026 & .920 & \multicolumn{1}{c|}{.955} & .049 & \textbf{.788} & \textbf{.877}\\
                \multicolumn{1}{c|}{$\mathrm{ICON_{22}}$ \cite{zhuge2022salient}} & \multicolumn{1}{c|}{30.09} & \multicolumn{1}{c|}{20.91} & .032 & .928 & \multicolumn{1}{c|}{.954} & .064 & .838 & \multicolumn{1}{c|}{.899} & .037 & .853 & \multicolumn{1}{c|}{.924} & .029 & .912 & \multicolumn{1}{c|}{.953} & .057 & .779 & .876\\
                \multicolumn{1}{c|}{$\mathrm{M^{3}Net_{23}}$ \cite{yuan2023m}} & \multicolumn{1}{c|}{34.61} & \multicolumn{1}{c|}{18.83} & \textbf{.029} & .926 & \multicolumn{1}{c|}{\textbf{.955}} & \textbf{.060} & .844 & \multicolumn{1}{c|}{\textbf{.904}} & \textbf{.037} & \textbf{.863} & \multicolumn{1}{c|}{\textbf{.927}} & \textbf{.026} & \textbf{.920} & \multicolumn{1}{c|}{\textbf{.959}} & .061 & .784 & .871\\
                \cline{1-18}
                \multicolumn{18}{c}{Lightweight CNN Models} \\
                \cline{1-18}
                 \multicolumn{1}{c|}{$\mathrm{HVPNet_{21}}$ \cite{liu2020lightweight}} & \multicolumn{1}{c|}{\textbf{1.24}} & \multicolumn{1}{c|}{1.1} & .052 & .882 & \multicolumn{1}{c|}{.911} & .089 & .783 & \multicolumn{1}{c|}{.844} & .058 & .772 & \multicolumn{1}{c|}{.859} & .044 & .867 & \multicolumn{1}{c|}{.913} & .065 & .736 & .839\\
                \multicolumn{1}{c|}{$\mathrm{SAMNet_{21}}$ \cite{liu2021samnet}} & \multicolumn{1}{c|}{1.33} & \multicolumn{1}{c|}{\textbf{0.5}} & .050 & .883 & \multicolumn{1}{c|}{.916} & .092 & .777 & \multicolumn{1}{c|}{.838} & .058 & .768 & \multicolumn{1}{c|}{.859} & .045 & .864 & \multicolumn{1}{c|}{.911} & .065 & .734 & .840\\
                \multicolumn{1}{c|}{$\mathrm{CII18_{21}}$ \cite{liu2021rethinking}} & \multicolumn{1}{c|}{11.89} & \multicolumn{1}{c|}{8.48} & .039 & .913 & \multicolumn{1}{c|}{.939} & .068 & .824 & \multicolumn{1}{c|}{.888} & .043 & .831 & \multicolumn{1}{c|}{.904} & .032 & .904 & \multicolumn{1}{c|}{.945} & .058 & .747 & .849\\
                \multicolumn{1}{c|}{$\mathrm{EDN(Lite)_{22}}$ \cite{wu2022edn}} & \multicolumn{1}{c|}{1.80} & \multicolumn{1}{c|}{1.14} & .042 & .910 & \multicolumn{1}{c|}{.933} & .073 & .818 & \multicolumn{1}{c|}{.877} & .045 & .819 & \multicolumn{1}{c|}{.895} & .034 & .897 & \multicolumn{1}{c|}{.937} & .057 & .746 & .848\\
                \rowcolor{gray!25}
                \multicolumn{1}{c|}{$\mathrm{Ours(S)}$} & \multicolumn{1}{c|}{1.66} & \multicolumn{1}{c|}{1.02} & .044 & .901 & \multicolumn{1}{c|}{.932} & .067 & .820 & \multicolumn{1}{c|}{.888} & .041 & .822 & \multicolumn{1}{c|}{.910} & .032 & .897 & \multicolumn{1}{c|}{.945} & .055 & .750 & .862\\
                \rowcolor{gray!25}
                \multicolumn{1}{c|}{$\mathrm{Ours(M)}$} & \multicolumn{1}{c|}{4.64} & \multicolumn{1}{c|}{2.67} & .037 & \.914 & \multicolumn{1}{c|}{.944} & .064 & .830 & \multicolumn{1}{c|}{.898} & .038 & .839 & \multicolumn{1}{c|}{.922} & .030 & .903 & \multicolumn{1}{c|}{.951} & .052 & .764 & .871\\
                \rowcolor{gray!25}
                \multicolumn{1}{c|}{$\mathrm{Ours(L)}$} & \multicolumn{1}{c|}{11.36} & \multicolumn{1}{c|}{3.25} & \textbf{.033} & \textbf{.921} & \multicolumn{1}{c|}{\textbf{.951}} & \textbf{.061} & \textbf{.836} & \multicolumn{1}{c|}{\textbf{.905}} & \textbf{.034} & \textbf{.853} & \multicolumn{1}{c|}{\textbf{.931}} & \textbf{.028} & \textbf{.908} & \multicolumn{1}{c|}{\textbf{.954}} & \textbf{.048} & \textbf{.782} & \textbf{.883}\\
                \cline{1-18}
    	\end{tabular}
	}
	\label{quan_result}
\end{table*}

The F-measure curves highlight that Ours (L) consistently maintains high scores across all thresholds, underlining its robust performance, while other models like HVPNet and EDN-Lite show a steeper decline in F-measure as the threshold increases, indicating reduced performance at higher strictness levels. This analysis not only benchmarks the capabilities of these models but also underscores the advancements in computational efficiency and algorithmic precision in the field of visual perception.

\section{Conclusions and Future Work}
In this paper, we introduced the Crossed Post-decoder Refinement (CPDR) architecture, a new approach to salient object detection that significantly enhances the performance of traditional models like U-Net and Feature Pyramid Networks. By incorporating a lightweight post-decoder refinement module that utilizes cross-level feature aggregation with both channel and spatial attention mechanisms, our model achieves superior saliency detection capabilities while maintaining reduced computational complexity. The CPDR model not only excels in benchmark tests against state-of-the-art methods but also offers a practical solution for deployment in environments with limited computational resources. For future work, an essential focus will be on optimizing the CPDR architecture for handling high-resolution images effectively while maintaining both accuracy and efficiency, since the increase of image resolution at the user-end results from the rapid development of imaging technology. This involves exploring strategies such as advanced down-sampling techniques to manage the increased computational load without losing significant image details crucial for accurate detection. Additionally, we are planning to further validate the universality of our approach, aiming to bring researchers a new plug-to-play module that is suitable for not only salient object detection but other widely studied computer vision tasks as well.

\section*{Acknowlegdgement}
The authors acknowledge the AI in Multimedia –  Image and Video Processing Laboratory (AIM-IVPL) for computing support and advice.

\bibliography{egbib}
\end{document}